\documentclass[runningheads]{llncs}
\usepackage[T1]{fontenc}
\usepackage{graphicx}
\usepackage{amssymb}
\usepackage{mathdots}
\usepackage{stmaryrd}
\usepackage{amsmath}
\usepackage{booktabs}
\usepackage{longtable}
\usepackage{xcolor,colortbl}
\usepackage{multirow, colortbl, xcolor, booktabs}
\usepackage{soul}
\usepackage{bm}
\usepackage{marvosym}

\definecolor{airforceblue}{rgb}{0.36, 0.54, 0.66}

\begin{document}
\title{Gene-induced Multimodal Pre-training for Image-omic Classification}
\titlerunning{Gene-induced Multimodal Pre-training for Image-omic Classification}

\author{Ting Jin\textsuperscript{1} \and Xingran Xie\textsuperscript{1} \and Renjie Wan\textsuperscript{2} \and Qingli Li\textsuperscript{1} \and
Yan Wang\textsuperscript{1}\textsuperscript{(\Letter)}}

\authorrunning{T. Jin et al.}
\institute{{\textsuperscript{1} Shanghai Key Laboratory of Multidimensional Information Processing, \\East China Normal University, Shanghai 200241, China\\\textsuperscript{2} Hong Kong Baptist University}\\
\email{51255904073@stu.ecnu.edu.cn,}
\email{xrxie@stu.ecnu.edu.cn,}
\email{wanpeoplejie@gmail.com,}
\email{qlli@cs.ecnu.edu.cn, ywang@cee.ecnu.edu.cn}}
\maketitle              
\begin{abstract}
Histology analysis of the tumor micro-environment integrated with genomic assays is the gold standard for most cancers in modern medicine. This paper proposes a Gene-induced Multimodal Pre-training (GiMP) framework, which jointly incorporates genomics and Whole Slide Images (WSIs) for classification tasks.  Our work aims at dealing with the main challenges of multi-modality image-omic classification \emph{w.r.t}. (1) the patient-level feature extraction difficulties from gigapixel WSIs and 
tens of thousands of genes, and (2) effective fusion considering high-order relevance modeling. Concretely, we first propose a group multi-head self-attention gene encoder to capture global structured features in gene expression cohorts. We design a masked patch modeling paradigm (MPM) to capture the latent pathological characteristics of different tissues. The mask strategy is randomly masking a fixed-length contiguous subsequence of patch embeddings of a WSI. Finally, we combine the classification tokens of paired modalities and propose a triplet learning module to learn high-order relevance and discriminative patient-level information. After pre-training, a simple fine-tuning can be adopted to obtain the classification results. Experimental results on the TCGA dataset show the superiority of our network architectures and our pre-training framework, achieving 99.47\% in accuracy for image-omic classification. The code is publicly available at https://github.com/huangwudiduan/GIMP.

\keywords{Multimodal learning\and Whole slide image classification }
\end{abstract}
%
%
%
\section{Introduction}
Pathological image-omic analysis is the cornerstone of modern medicine and demonstrates promise in a variety of different tasks such as cancer diagnosis and prognosis \cite{WHO}. With the recent advance of digital pathology and sequencing technologies, modern cancer screening has jointly incorporated genomics and histology analysis of whole slide images (WSIs).

Though deep learning techniques have revolutionized medical imaging, designing a task-specific algorithm for image-omic multi-modality analysis is challenging. (1) The gigapixel WSIs, which generally yield 15,000 foreground patches during pre-processing, make attention-based backbones \cite{abmil} hard to extract precise image (WSI)-level representations. (2) Learning features from genomics data which have tens of thousands of genes make models such as Transformer \cite{selfattn} impractical to use due to its quadratic computation complexity. (3) Image-omic feature fusion \cite{pathomicfusion,mcat} may fail to model high-order relevance and the inherent structural characteristics of each modality, making the fusion less effective.

Specifically, to our knowledge, most multi-modality techniques have been designed for modalities such as chest X-ray and reports \cite{BilViL,mgca,refers}, CT and X-ray \cite{Xie2022unimiss}, CT and MRI \cite{Yang2022toward}, H\&E cross-staining \cite{csco} via global feature, local feature or multi-granularity alignment. But, none of these works considers the challenges in WSIs and genes processing. Besides, vision-language models in the computer vision community stand out for their remarkable versatility \cite{clip,texttoimg}. Nevertheless, constrained by computing resources, the most commonly used multimodal representation learning strategy, contrastive learning, which relies on a large number of negative samples to avoid model collapse \cite{mutual}, is not affordable for gigapixel WSIs analysis. A big domain gap also hampers their usage in leveraging the structural characteristic of tumor micro-environment and genomic assay. Recently, the literature corpus has proposed some methods for accomplishing specific image-omic tasks via Kronecker Product fusion \cite{pathomicfusion} or co-attention mapping between WSIs and genomics data \cite{mcat}. But, the Kronecker product overly concerns feature interactions between modalities while ignoring high-order relevance, \emph{w.r.t.} decision boundaries across multiple samples, which is critical to classification tasks. As for the co-attention module, it is unidirectional and cannot localize significant regions from genetic data with a large amount of information. 

In this paper, we propose a task-specific framework dubbed Gene-induced Multimodal Pre-training (GiMP) for image-omic classification. Concretely, we first propose a transformer-based gene encoder, Group Multi-head Self Attention (GroupMSA), to capture global structured features in gene expression cohorts. Next, we design a pre-training paradigm for WSIs, Masked Patch Modeling (MPM), masking random patch embeddings from a fixed-length contiguous subsequence of a WSI. We assume that one patch-level feature embedding can be reconstructed by its adjacent patches, and this process enhances the learning ability for pathological characteristics of different tissues. Our MPM only needs to recover the masked patch embeddings in a fixed-length subsequence rather than processing all patches from WSIs. Furthermore, to model the high-order relevance of the two modalities, we combine CLS tokens of paired image and genomic data to form unified representations and propose a triplet learning module to differentiate patient-level positive and negative samples in a mini-batch. It is worth mentioning that although our unified representation fuses features from the whole gene expression cohort and partial WSIs in a mini-batch, we can still learn high-order relevance and discriminative patient-level information between these two modalities in pre-training thanks to the triplet learning module. In addition, note that our proposed method is different from self-supervised pre-training. Specifically, we focus not only on superior representation learning capability, but also category-related feature distributions, \emph{w.r.t.} intra- and inter-class variation. With the training process going on, complete information from WSIs can be integrated and the fused multimodal representations with high discrimination will make it easier for the classifier to find the classification hyperplane. Experimental results demonstrate that our GiMP achieves significant improvement in accuracy than other image-omic competitors, and our multimodal framework shows competitive performance even without pre-training.

\begin{figure}[!tb]
\begin{center}
\includegraphics[width=0.9\textwidth]{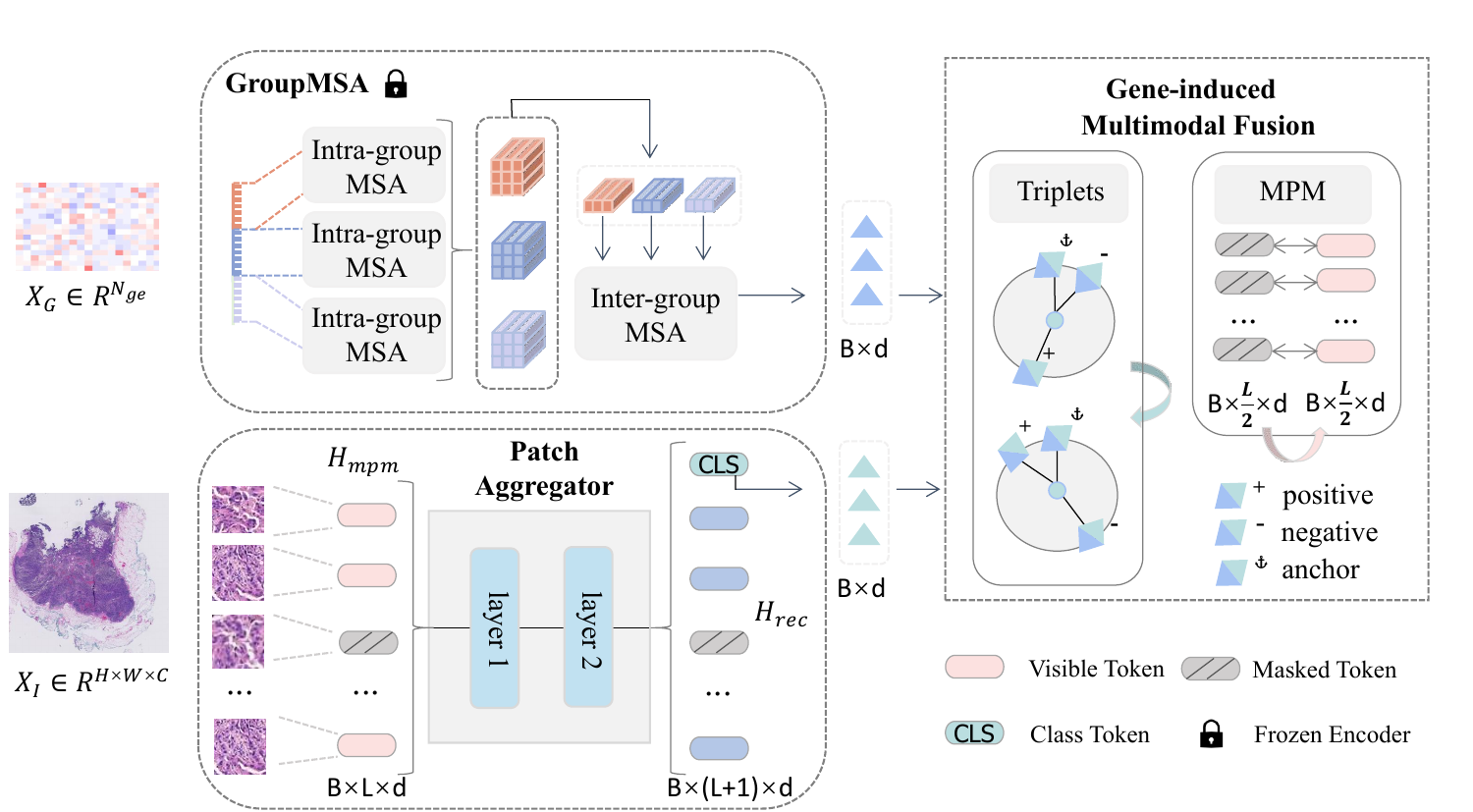}
\end{center}
\caption{Illustration of GiMP pre-training. Given a batch of image-omic pairs, we randomly select a fixed-length patch cohort and mask parts of the patch embeddings. Then we use two modality-specific encoders to capture unimodal features. Two pre-training objectives are considered: 1) building triplets by concatenated CLS tokens of each modality and enhancing the discriminability according to category relations, and 2) reconstructing the missing patch embeddings by its adjacent patches.} \label{fig:framework}
\end{figure}

\section{Method}
Given a multimodal dataset $\mathcal{D}$ consisting of pairs of WSI pathological images and genomic data $(\mathbf{X}_I, \mathbf{X}_G)$, our GiMP learns feature representations via accomplishing masked patch modeling and triplets learning. As shown in Fig.~\ref{fig:framework}, the overall framework consists of three parts: 1) group-based genetic encoder GroupMSA (Sec.~\ref{sec:2.1}), 2) efficient patch aggregator (Sec.~\ref{sec:2.2}) and 3) gene-induced multimodal fusion (Sec.~\ref{sec:2.3}). In the subsequent sections, we will introduce each part of our proposed framework in detail.

\subsection{Group Multi-head Self Attention}\label{sec:2.1}
In this section, we propose Group Multi-head Self Attention (GroupMSA), a specialized gene encoder to capture structured features in genomic data cohorts. Specifically, inspired by tokenisation techniques in natural language processing \cite{selfattn}, the input expression cohort $\mathbf{X}_G\in\mathbb{R}^{N_{ge}}$ is partitioned into $N_f$  non-overlapping fragments, and we then use a linear projection head to acquire fragment features $\mathbf{H}_f\in\mathbb{R}^{N_f\times d}$, where $d$ is the hidden dimension. Next, we introduce an intra-and-inter attention module to capture local and global information in $\mathbf{H}_f$. Firstly, the fragment features are divided into groups and there are $N_{gr}$ learnable group tokens linked to each group resulting in $(N_f / N_{gr} + 1)$ tokens per group. Then the prepared tokens are fed to a vanilla multi-head self-attention (MSA) block to extract intra-group information. After that, we model cross-group interactions by another MSA layer on the global scale with the locally learned group tokens and a final classification token $\mathbf{CLS}_{ge}\in\mathbb{R}^{d}$. Finally, GroupMSA could learn dense semantics from the genomic data cohort.

\subsection{Patch Aggregator with Efficient Attention Operation}\label{sec:2.2}
Let's denote the whole slide pathological image with $H\times W$ spatial resolution and $C$ channels by $\mathbf{X}_I\in\mathbb{R}^{H \times W \times C}$. We follow the preprocessing strategy of CLAM \cite{clam} to acquire patch-level embedding sequence, \emph{i.e.}, each foreground patch with $256\times256$ pixels is fed into an ImageNet-pretrained ResNet50 and the background region is discarded. Let $\mathbf{H}_p = \left\{h_{j} \mid h_{j} \in \mathbb{R}^{1024}\right\}_{j=1}^{N_p}$ denote the sequence of patch embeddings corresponding to WSI $\mathbf{X}_I$ and note that the total patch number $N_p$ is image-specific. Since the quadratic computational complexity of the standard self-attention mechanism is usually unaffordable in WSI analysis due to its long instances sequence, we employ Nystrom-based attention algorithm \cite{nystromformer} to aggregate patch embeddings and yield image-level predictions. Specifically, the input sequence $\mathbf{H}_p$ is first embedded into a $d$-dimensional feature space and combined with a classification token $\mathbf{CLS}_{img}$, yielding $\mathbf{H}^{0}_p \in \mathbb{R}^{\left(N_p+1\right) \times d}$. Then we perform different projection operations on $\mathbf{H}^{0}_p$:
\begin{equation}
\label{eq:qkv}
    \mathbf{Q}^l=\mathbf{H}_p^l\cdot W_Q^l, \mathbf{K}^l=\mathbf{H}_p^l\cdot W_K^l, \mathbf{V}^l=\mathbf{H}_p^l\cdot W_V^l,
\end{equation}
\begin{equation}
\label{eq:nystromattn}
    \mathbf{H}_p^{l+1} = \operatorname{softmax}(\frac{\mathbf{Q}^l\cdot \widetilde{\mathbf{K}}^{l\top}}{\sqrt{d}})\cdot\left(\operatorname{softmax}(\frac{\widetilde{\mathbf{Q}}^l\cdot\widetilde{\mathbf{K}}^{l\top}}{\sqrt{d}})\right)^{-1} \cdot\operatorname{softmax}(\frac{\widetilde{\mathbf{Q}}^l\cdot\mathbf{K}^{l\top}}{\sqrt{d}})\cdot\mathbf{V}^l,
\end{equation}
where $W_Q^l, W_K^l, W_V^l \in\mathbb{R}^{d\times d}$ are linear mapping matrices, $\widetilde{\mathbf{Q}}^l, \widetilde{\mathbf{K}}^l\in\mathbb{R}^{m\times d}$ ($m\ll N_p$) are downsampling matrices obtained from clustering tokens in $\mathbf{Q}^l$ and $\mathbf{K}^l$ for layer $l\in\{0,1\}$.

\subsection{Gene-induced Multimodal Fusion}\label{sec:2.3}
In this section, we first describe the formulation of masked patch modeling. Then we introduce the overall pipeline of our pre-training framework and illustrate how to apply it to downstream classification tasks.
\subsubsection{Masked Patch Modeling}
In WSIs, the foreground patches are spatially contiguous, which means the adjacent patches have similar feature embeddings. Thus, we propose a Masked Patch Modeling (MPM) pre-training strategy that masks random patch embeddings from a fixed-length contiguous subsequence $\mathbf{H}_{mpm}=\left\{h_{j} \mid h_{j}\in\mathbb{R}^{1024}\right\}_{j=i}^{L+i}$ in $\mathbf{H}_p$ and reconstruct the invisible information. The fixed subsequence length $L$ is empirically set to 6,000 and the sequences shorter than $L$ are duplicated to build mini batches. Besides, the masking ratio is set to $50\%$ and the set of masked subscripts is denoted as $\mathcal{M}\in\mathbb{R}^{0.5L}$. Next, a two-layer Nystrom-based patch aggregator followed by a lightweight reconstruction decoder are adopted to process the masked sequence $\mathbf{H}_{mpm}$ and the reconstructed sequence is denoted as $\mathbf{H}_{rec}=\left\{\hat{h}_{j}\mid\hat{h}_{j}\in\mathbb{R}^{1024}\right\}_{j=1}^L$. Note that we reconstruct the missing feature embeddings rather than the raw pixels of the masked areas, which is different from traditional MIM methods like SimMIM \cite{simmim} and MAE \cite{mae}. In this way, the model could consider latent pathological characteristics of different tissues, which makes the pretext task more challenging. The reconstruction $L_1$ loss is computed by:
\begin{equation}
\label{eq:recloss}
    \mathcal{L}_{rec}=\sum_{j=1}^L\mathbf{1}[j\in \mathcal{M}]\left\|h_{j}-\hat{h}_{j}\right\|_1,
\end{equation}
where $\mathbf{1}[\cdot]$ is the indicator function.
\subsubsection{Gene-induced Triplet Learning}
The transformer-based backbones in the classification task require the CLS token to be able to extract accurate global information, which is even more important yet difficult in WSIs due to the long sequence challenge. In addition, in order to construct the mini-batch, the subsequences we intercept in the MPM pre-training phase may not be sufficiently representative of the image-level characteristics. To overcome these issues, we further propose a gene-induced triplet learning module, which uses pathological images and genomic data as input and extracts high-order and discriminative features via CLS tokens. Firstly, we pre-train the GroupMSA module by patient-level annotations in advance and froze it in the following iterations. Next, a learnable CLS token $\mathbf{CLS}_{img}$ for WSIs is added to the input masked sequence $\mathbf{H}_{mpm}$. After extracting the input patch embeddings and gene sequence separately, we concatenate $\mathbf{CLS}_{img}$ and $\mathbf{CLS}_{ge}$ as $\mathbf{CLS}_{pat}\in\mathbb{R}^{2d}$ to represent patient-level characteristics.

Suppose we obtain a triplet list $\left\{x, x^+, x^-\right\}$ during current iteration, where $x, x^+, x^-$ are concatenated tokens of anchor $\mathbf{CLS}_{pat}$, positive $\mathbf{CLS}_{pat}$, and negative $\mathbf{CLS}_{pat}$, respectively. To enhance the global modeling capability, \emph{i.e.}, extracting more precise patient-level features, we expect that the distance between the anchor and the positive sample gets closer, while the negative sample is farther away. The loss function for optimizing triplet learning is computed by:
\begin{equation}
\label{eq:tripletloss}
    \mathcal{L}_{tri}=\operatorname{max}(\left\|x-x^+\right\|_2^2+\delta-\left\|x-x^-\right\|_2^2, 0),
\end{equation}
$\delta$ indicates a threshold, \emph{e.g.}, $\delta=0.8$. Finally, the loss function for GiMP pre-training is: $\mathcal{L}_{pre}=\mathcal{L}_{tri}+\mathcal{L}_{rec}$.
\subsubsection{Multimodal Fine-tuning}
Applying the pre-trained backbone to image-omic classification task is straightforward, since GiMP pre-training allows it to learn representative patient-level features. We use a simple Multi-Layer Perceptron (MLP) head to map $\mathbf{CLS}_{pat}$ to the  final class predictions $\hat{P}$, which can be written as $\hat{P}=\operatorname{softmax}(\operatorname{MLP}(\mathbf{CLS}_{pat}))$.

\section{Experiments}

\subsection{Experimental Setup}\label{sec:3.1}
\subsubsection{Datasets} 
We verify the effectiveness of our method on The Caner Genome Atlas (TCGA) non-small cell lung cancer (NSCLC) dataset, which contains two cancer subtypes, \emph{i.e.}, Lung Squamous Cell Carcinoma (LUSC) and Lung Adenocarcinoma (LUAD). After pre-processing \cite{clam}, the patch number extracted from WSIs at 20× magnification varies from 485 to 148,569. We collect corresponding RNA-seq FPKM data for each patient and the length of the input genomic sequence is 60,480. Among 946 image-omic pairs, 470 of them belong to LUAD and 476 cases are LUSC. We randomly split the data into 567 for training, 189 for validation and 190 for testing. 

\subsubsection{Implementation Details}
The pre-training process of all algorithms is conducted on the training set, without any extra data augmentation. Note that our genetic encoder, GroupMSA, is fully supervised pre-trained on {unimodal} genetic data to accelerate convergence and it is frozen during GiMP training process. The maximum pre-training epoch for all methods is set to 100 and we fine-tune the models at the last epoch. During fine-tuning, we evaluate the model on the validation set after every epoch, and save the parameters when it performs the best. AdamW \cite{AdamW} is used as our optimizer and the learning rate is $10^{-4}$ with cosine decline strategy. The maximum number of fine-tune epoch is 70. At last, we measure the performance on the test set. Training configurations are consistent throughout the fine-tuning process to ensure fair comparisons. All experiments are conducted on a single NVIDIA GeForce RTX 3090.
 
\begin{table}[t]
\renewcommand\arraystretch{0.9}
\setlength\tabcolsep{3.5pt}
\footnotesize
\centering
\caption{Accuracy comparison on the TCGA Lung Cancer dataset. The best results are marked in \textbf{bold}.}
\label{tab:comparison}
\begin{tabular}{clclllllc}
\toprule[0.15em]
\textbf{Modality} & & \textbf{Pre-train} & & & \textbf{Method} & & & \textbf{Acc.} \\ \midrule[0.09em]
& & & & & ABMIL \cite{abmil} & & & 0.7737 \\
& & & & & DSMIL \cite{dsmil} & & & 0.7566 \\
& & & & & CLAM-SB \cite{clam} & & & 0.8519 \\
& & & & & CLAM-MB \cite{clam} & & & 0.8889 \\
& & & & & TransMIL \cite{transmil} & & & 0.8836 \\
\multirow{-6}{*}{Pathology} & & \multirow{-6}{*}{w/o pre-train} & & & \cellcolor[HTML]{EFEFEF}\textbf{GiMP (w/o GroupMSA)} & \cellcolor[HTML]{EFEFEF} & \cellcolor[HTML]{EFEFEF} & \cellcolor[HTML]{EFEFEF}\textbf{0.8995} \\ 
\midrule
& & & & & PORPOISE \cite{porpoise} & & & 0.9524 \\
& & & & & Pathomic Fusion \cite{pathomicfusion} & & & 0.9684 \\
& & & & & MCAT \cite{mcat} & & & 0.9632 \\
& & \multirow{-4}{*}{w/o pre-train} & & & \cellcolor[HTML]{EFEFEF}\textbf{GiMP (ours)} & \cellcolor[HTML]{EFEFEF} & \cellcolor[HTML]{EFEFEF} & \cellcolor[HTML]{EFEFEF}\textbf{0.9737} \\ \cline{6-9}
& & & & & MGCA \cite{mgca} & & & 0.9105 \\
& & & & & BioViL \cite{BilViL} & & & 0.9316 \\
& & & & & REFERS \cite{refers} & & & 0.9368 \\
\multirow{-8}{*}{\begin{tabular}[c]{@{}c@{}}Pathology\\ \&\\ Genomic\end{tabular}} & & \multirow{-4}{*}{w/ pre-train} & & & \cellcolor[HTML]{EFEFEF}\textbf{GiMP (ours)} & \cellcolor[HTML]{EFEFEF} & \cellcolor[HTML]{EFEFEF} & \cellcolor[HTML]{EFEFEF}\textbf{0.9947} \\
\bottomrule[0.15em]
\end{tabular}
\end{table}

\subsection{Comparison between GiMP and Other Methods \label{sec:3.2}}
We conduct comparisons between GiMP and three competitors under different settings. Firstly, we compare our proposed patch aggregator with the current state-of-the-art deep MIL models on \emph{unimodal} TCGA-NSCLC dataset, \emph{i.e.}, only pathological WSIs are included as input. As shown in Table~\ref{tab:comparison}, our proposed patch aggregator outperforms all the compared attention based multiple instance learning baselines in classification accuracy. In particular, 1.6$\%$ higher than the second best compared method TransMIL \cite{transmil}. We then explore the superiority of GiMP by comparing to state-of-the-art medical multi-modal approaches. We particularly compare our method to BioViL \cite{BilViL}, MGCA \cite{mgca} and REFERS \cite{refers}, three popular multimodal pre-training algorithms in medical text-image classification task. We can observe in the table that, our GiMP raises ACC from 91.05$\%$ to 99.47$\%$ on TCGA-NSCLC dataset. Even without pre-training stage, GiMP shows competitive performance compared to PORPOISE \cite{porpoise}, Pathomic Fusion \cite{pathomicfusion}, and MCAT \cite{mcat}, three influential image-omic classification architectures. 

\begin{figure}[!tb]
\begin{center}
\includegraphics[width=0.95\textwidth]{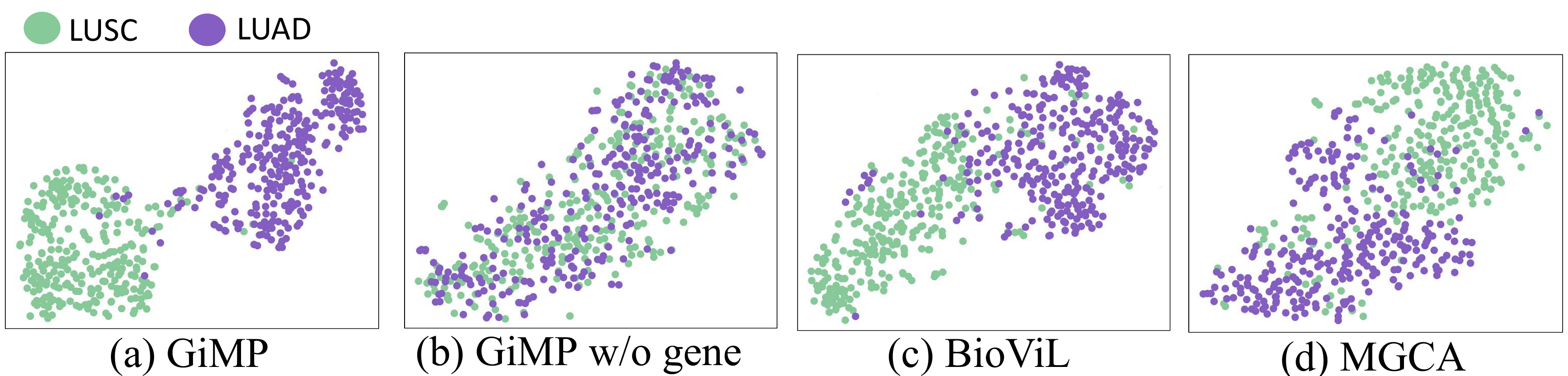}
\end{center}
\caption{t-SNE visualization of different methods with $\mathbf{CLS}_{pat}$ after pre-training. (a) image-omic GiMP pre-trained, (b) GiMP pre-trained without gene inducing, (c) BioViL \cite{BilViL} pre-trained, (d) MGCA \cite{mgca} pre-trained.} \label{fig:tsne}
\end{figure}

\begin{table}[t]
\renewcommand\arraystretch{0.85}
\setlength{\tabcolsep}{4pt}
\footnotesize
\centering
\caption{Ablation study on TCGA Lung Cancer dataset. "SNN" means replacing GroupMSA with SNN \cite{snn}. "Triplet" denotes our gene-induced triplet learning module.}
\label{tab:ablation}
\begin{tabular}{cccc|c}
\hline
\toprule[0.15em]
\textbf{Aggregator} & \textbf{GroupMSA} & \textbf{Triplet} & \textbf{MPM} & \textbf{Acc.}\\ 
\midrule[0.09em]
\checkmark & SNN \cite{snn} & & & 0.9684\\
\checkmark & \checkmark & & & 0.9737\\
\checkmark & & & \checkmark & 0.9579\\
\checkmark & & \checkmark & & 0.9263\\
\checkmark & \checkmark & \checkmark & & 0.9526 \\
\rowcolor[HTML]{EFEFEF} 
\checkmark & \checkmark & \checkmark & \checkmark & \textbf{0.9974} \\ 
\bottomrule[0.15em]
\end{tabular}
\end{table}

We further explore why GiMP works by insightful interpretation of the proposed method with t-SNE visualisation. Fig.~\ref{fig:tsne} shows the feature mixtureness of pre-trained $\mathbf{CLS}_{pat}$ extracting global information on training set. Comparison between Fig.~\ref{fig:tsne} (a) and (b) indicates that the addition of the genomic data is indispensable in increasing the inter-class distance and reducing the intra-class distance, which confirms our motivation that gene-induced multimodal fusion could model high-order relevance and yield more discriminative representations. Moreover, compared to the mentioned self-supervised methods BioViL \cite{BilViL} and MGCA \cite{mgca} in Fig.~\ref{fig:tsne} (c) and (d), $\mathbf{CLS}_{pat}$ with GiMP pre-trained are well separated between LUAD and LUSC, \emph{i.e.}, GiMP pays more attention to the category-related feature distribution and could extract more discriminative patient-level features during triplet learning. 

\subsection{Ablation Study\label{sec:3.3}}
Table~\ref{tab:ablation} summarizes the results of ablation study. We first evaluate the effectiveness of the proposed GroupMSA. In the first two rows, GroupMSA achieves 0.53\% improvement compared to SNN \cite{snn}, a popular genetic encoders used in PORPOISE \cite{porpoise} and Pathomic Fusion \cite{pathomicfusion}. We then analyze the effect of adding genetic modality during pre-training. The evaluation protocol is first pre-training, and then fine-tuning on downstream multimodal classification task.  “Aggregator + MPM” means GiMP only uses WSIs as input and reconstructs the missing patch embeddings during the pre-training phase. Since the fixed subsequence length $L=6000$ is used in our setting, it is sometimes smaller than the original patch number, \emph{e.g.}, the maximum size 148,569, the pre-trained model without genetic guidance may be not aware of sufficiently accurate patient-level characteristics, \emph{i.e.}, ineffectively focused on normal tissues. “Aggregator + Triplet” indicates using unimodal image features to build triplets. We can likewise find that the lack of precise global representation leads to worse performance. 
Finally, we evaluate the necessity of the MPM module. “Aggregator + GroupMSA + Triplet” means GiMP only combines the CLS tokens of each modality and calculates triplet loss during pre-training. We can observe a performance drop without MPM module, \emph{e.g.}, from 99.47\% to 95.26\%, which demonstrates that local pathological information is equally critical as high-order relevance.

\section{Conclusion}
In this paper, we propose a novel multimodal pre-training method to exploit the complementary relationship of genomic data and pathological images. Concretely, we introduce a genetic encoder with structured learning capabilities and an effective gene-induced multimodal fusion module which combines two pre-training objectives, triplet learning and masked patch modeling. Experimental results demonstrate the superior performance of the proposed GiMP compared to other state-of-the-art methods. The contribution of each proposed component of GiMP is also demonstrated in the experiments. 

\subsubsection{Acknowledgements} This work was supported by the National Natural Science Foundation of China (Grant No. 62101191), Shanghai Natural Science Foundation (Grant No. 21ZR1420800), and the Science and Technology Commission of Shanghai Municipality (Grant No. 22DZ2229004).


%
%

\bibliographystyle{splncs04}
\bibliography{mybib.bib}
\end{document}


\title{Supplementary Material for ``Gene-induced Multimodal Pre-training for Image-omic Classification''}
\titlerunning{Gene-induced Multimodal Pre-training for Image-omic Classification}
\author{Ting Jin\textsuperscript{1} \and Xingran Xie\textsuperscript{1} \and Renjie Wan\textsuperscript{2} \and Qingli Li\textsuperscript{1} \and
Yan Wang\textsuperscript{1}\textsuperscript{(\Letter)}}

\authorrunning{T. Jin et al.}
\institute{{\textsuperscript{1} Shanghai Key Laboratory of Multidimensional Information Processing, \\East China Normal University, Shanghai 200241, China\\\textsuperscript{2} Hong Kong Baptist University}\\
\email{51255904073@stu.ecnu.edu.cn,}
\email{xrxie@stu.ecnu.edu.cn,}
\email{wanpeoplejie@gmail.com,}
\email{qlli@cs.ecnu.edu.cn, ywang@cee.ecnu.edu.cn}}
%
%
\maketitle


\begin{figure}[]
\begin{center}
\includegraphics[width=0.95\textwidth]{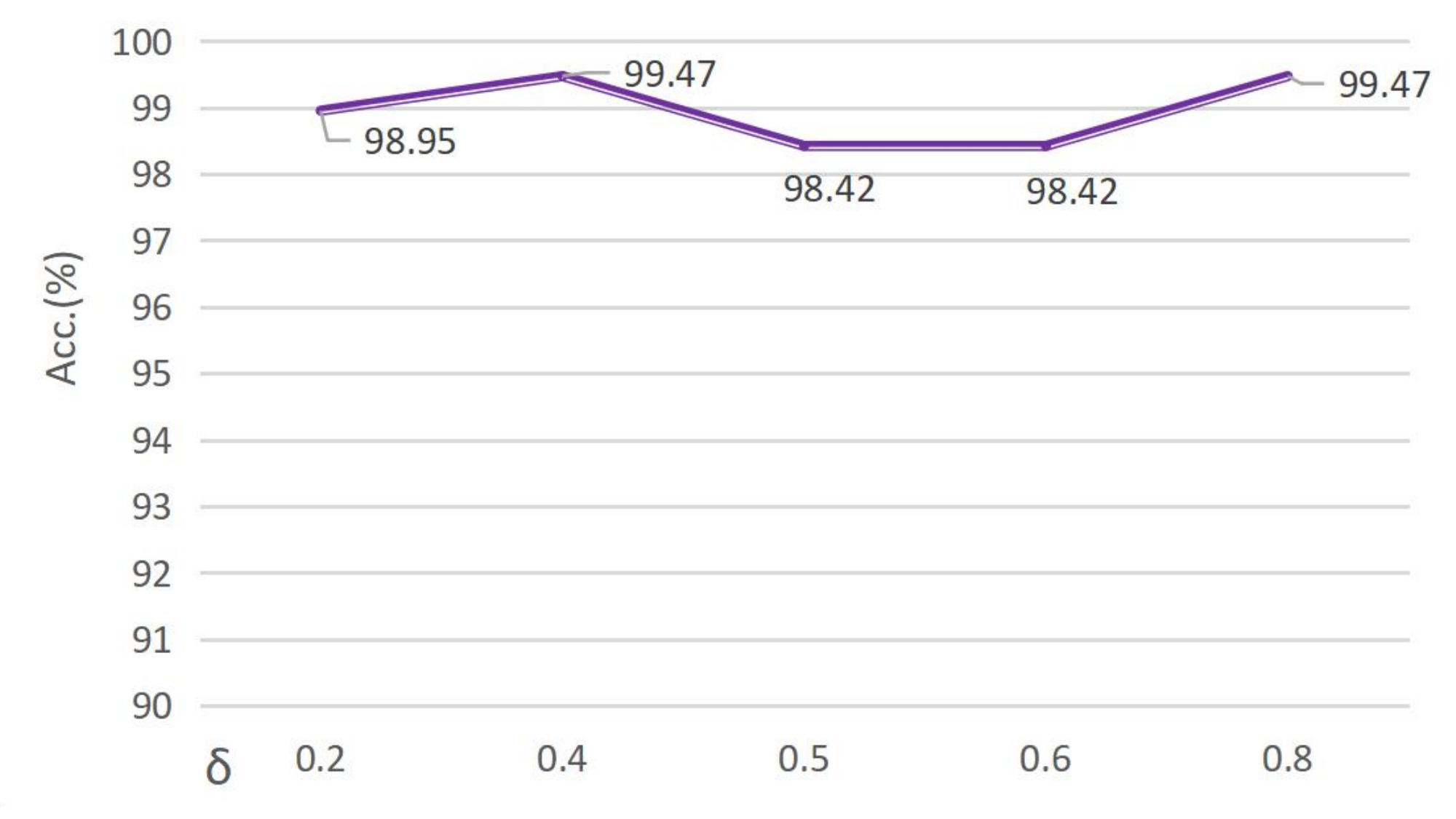}
\end{center}
\caption{Parameter Analysis. $\delta$ indicates the threshold in Eq. 4 for optimizing triplet learning. $\delta$ is not sensitive within the range of $[0.2, 0.8]$.} \label{fig:tsne}
\end{figure}
